\documentclass[10pt,twocolumn,letterpaper]{article}

\usepackage{cvpr}
\usepackage{times}
\usepackage{epsfig}
\usepackage{graphicx}
\usepackage{amsmath}
\usepackage{amssymb}
\usepackage{diagbox}
\usepackage{multirow}
\usepackage{tablefootnote}

\usepackage[pagebackref=true,breaklinks=true,letterpaper=true,colorlinks,bookmarks=false]{hyperref}

\cvprfinalcopy 

\def\cvprPaperID{73} 

\newcommand{\tabincell}[2]{\begin{tabular}{@{}#1@{}}#2\end{tabular}}

\ifcvprfinal\fi
\begin{document}

\title{Learning Face Representation from Scratch}

\input{latex.macro}
\graphicspath{{images/}}

\author{Dong Yi, Zhen Lei, Shengcai Liao and Stan Z. Li\\
Center for Biometrics and Security Research \& National Laboratory of Pattern Recognition\\
Institute of Automation, Chinese Academy of Sciences (CASIA)\\
{\tt\small dong.yi, zlei, scliao, szli@nlpr.ia.ac.cn}
}

\maketitle

\begin{abstract}
Pushing by big data and deep convolutional neural network (CNN), the performance of face recognition is becoming comparable to human. Using private large scale training datasets, several groups achieve very high performance on LFW, \ie 97\% to 99\%. While there are many open source implementations of CNN, none of large scale face dataset is publicly available. The current situation in the field of face recognition is that data is more important than algorithm. To solve this problem, this paper proposes a semi-automatical way to collect face images from Internet and builds a large scale dataset containing about 10,000 subjects and 500,000 images, called CASIA-WebFace. Based on the database, we use a 11-layer CNN to learn discriminative representation and obtain state-of-the-art accuracy on LFW and YTF. The publication of CASIA-WebFace will attract more research groups entering this field and accelerate the development of face recognition in the wild.
\end{abstract}

\section{Introduction}

In the past year, the performance of face recognition algorithms increased in a large margin. For example, the accuracy on LFW~\cite{Huang-LFW-2007}, the hardest face dataset at present, is improved from 95\% to 99\%~\cite{Sun-NIPS-2014}, which is on a par with human performance. The best methods on LFW can be divided into two categories: wide model and deep model. The essence of good model is that it should has enough capacity to represent the variations of complex face images. High dimensional LBP~\cite{Chen-CVPR-2013} is a typical wide model which flatten the face manifold by transforming the image into a very high dimensional space. And convolutional neural network (CNN)~\cite{Lecun-IEEE-1998} is the state-of-the-art deep model for face recognition and image analysis.

Although a model can increase its complexity either along ``width'' or ``depth'' direction, deep model is more effective than wide model when with the same number of parameters~\cite{Bengio-AI-2009}. Furthermore, ordinary computer can't easily handle the high dimensional features extracted by wide model. On the contrary, the dimension of features in each layer of deep model are much lower, which make the memory consumption of deep model is affordable. The power of CNN to distill knowledge from data has been verified in may fields~\cite{Krizhevsky-NIPS-2012}. Recently, deep CNN is becoming the mainstream in face recognition and hold the top positions on LFW. However, limited by the scale of training data, the ceiling of CNN has not been measured yet.

Many excellent open source implementations~\cite{cuda-convnet,caffe} of CNN can be used to learn face representations from big data but no research team has made their private large dataset public until now. The reason may be that large datasets are very hard to collect and need consume lots of money and manpower. But for academy research, developing algorithms on private data is harmful in two aspects: First, most researchers can't make contributions to large scale face recognition methods for lacking of data. Second, due to different training set, many classical methods and CNN are not comparable.

On LFW, the best methods both use outside data besides of LFW, \ie the ``Unrestricted, Labeled Outside Data'' category of LFW. In fact, the methods in this category are hardly called as methods simply but solutions which at least include outside data and algorithm. To extend the scale of LFW and standardize the evaluation protocol of ``Unrestricted, Labeled Outside Data'', this paper builds a large scale dataset including about 10,000 subjects and 500,000 images, called CASIA-WebFace~\footnote{You could apply the dataset at \url{http://www.cbsr.ia.ac.cn/english/CASIA-WebFace-Database.html}}. To the best of our knowledge, the size of this dataset rank second in the literature, only smaller than the private dataset of Facebook (SCF)~\cite{Taigman-CVPR-2014}. We encourage those data-consuming methods training on this dataset and reporting performance on LFW.

Crawling face images from Internet is easy but annotating their identities is hard. Thanks to the good structure of IMDb~\footnote{\url{http://www.imdb.com}} website, the crawling and annotation can be done in a semi-automatic way. First, the names of some interested celebrities are crawled form the website, and then the photos in their pages are downloaded. Because most photos usually contain more than one face, the difficult arise. Thus we propose a simple and fast clustering method to annotate the identity of faces in the photos. To ensure the subjects in the dataset are not overlapping to LFW, we use edit distance of the names to check duplication. Finally, we scan the whole dataset by manual and correct the false annotations.

To illustrate the quality of CASIA-WebFace, we train a deep CNN on it. Our network integrates the most popular components in the recent works, such as ReLU neuron~\cite{Nair-ICML-2010}, dropout~\cite{Hinton-Dropout-2012}, low dimensional representation, identification + verification cost function~\cite{Sun-NIPS-2014},  small filter and very deep architecture~\cite{Simonyan-2014}. Our network is evaluated on LFW according to the standard protocol and a newly proposed protocol ``BLUFER''~\cite{Liao-IJCB-2014}. All experimental results show its superior performance. The network is also tested on another challenging dataset, YouTube Faces (YTF)~\cite{Wolf-CVPR-2011}. Using the representations learned from CASIA-WebFace, we achieve comparable result to Facebook's DeepFace~\cite{Taigman-CVPR-2014}.

The contributions of this paper are summarized as follows
\begin{itemize}
\item We build a large scale face dataset and makes it public, which will dispel the chaos of evaluation on LFW and make the methods fairly comparable;
\item We propose a semi-automatic pipeline to construct large scale face dataset from Internet, which will attract more researchers to build new face datasets or enlarge existing face datasets;
\item We train a high performance baseline deep CNN for face recognition in the wild. When using {\bf Single} network, the performance of our network is better than DeepFace~\cite{Taigman-CVPR-2014} and DeepID2~\cite{Sun-NIPS-2014}.
\end{itemize}

\section{Related Work}
\label{sec:review}

Data and algorithm are two essential components for pattern recognition. With the successful applications of deep learning in face recognition, dataset collection lags behind algorithm. In this section we review some popular face datasets and representation learning methods.

\subsection{Face Dataset}


Early face datasets were almost collected under controlled environments, such as PIE~\cite{Sim-CMU-2002}, FERET~\cite{Phillips-FERET-PAMI-00} and so on. Through the efforts of many researchers, we could obtain very high performance on these ideal datasets. But we found that the models learned from these datasets are difficult to generalize to new environments in practical applications. To improve the generalization of face recognition methods, the interests of community gradually transferred from controlled environments to uncontrolled environments, \ie face recognition in the wild. Then a milestone dataset, LFW~\cite{Huang-LFW-2007} including 5749 subjects, was born in 2007.

Compared to previous datasets, the biggest difference of LFW is that the images are crawled from Internet rather than acquired under several pre-defined environments. Therefore, LFW has more variations in pose, illumination, expression, resolution, imaging device and these factors are combined together in random way. In 2009, based on the name list of LFW \cite{Kumar-ICCV-2009} collected another good dataset named as PubFig. Although PubFig just include 200 subjects, the number of image for each subject is much more than LFW and it supply 73 attributes to describe the face images. YTF~\cite{Wolf-CVPR-2011} is another dataset based on the name list of LFW but it's created for video based face recognition. All the videos in YTF were downloaded from YouTube. Because the videos on YouTube are compressed in very high ratio, the quality of the face snapshots are lower than LFW.

CACD~\cite{Chen-ECCV-2014} is a large dataset collected for cross-age face recognition in 2014, which include 2,000 subjects and 163,446 images. The scale of CACD is large enough to train deep models but the dataset contains much noise and incorrect identity labels. The reason is that the images are crawled by Google Image search engine, and just a small subset (200 subjects) is checked by manual.

Besides the above publicly accessible datasets, there are three large scale private datasets: Facebook's SFC~\cite{Taigman-CVPR-2014}, CUHK's CelebFaces~\cite{Sun-NIPS-2014} and MSRA's WDRef~\cite{Chen-ECCV-2012}. Among them, SFC has the biggest scale, including more then 4000 subjects and each subject has an average of 1000 images. Using SFC, \cite{Taigman-CVPR-2014} successfully learns an effective face representation robust to face variations in the wild. Although the scale of CelebFaces and WDRef are relative smaller than SFC, they are also good resources for developing high performance algorithms. The current state-of-the-art accuracy on LFW is obtained by training on CelebFaces. It's a pity that the three good datasets are not publicly available, therefore this paper collects CASIA-WebFace to fill this gap.

\subsection{Face Representation Learning}

The first popular face recognition method is Eigenface~\cite{Turk-91} which was proposed in 1991. Now we can see Eigenface as a model with one linear layer. Fisherface~\cite{Belhumeur-CVPR-96} or LDA is a one layer linear model too. In the following long period, researchers mainly focused on how to solve the parameters of the linear layer with respect to some cost functions, such as reconstruction error and classification error. Much attentions were also paid to regularized the solution of LDA because LDA is easily prone to SSS problem (Small Sample Size).

Then, various local feature based methods emerged and they were naturally used by combining with the above linear models, such as Gabor+LDA~\cite{Liu-Gabor-2002}, LBP+LDA~\cite{Li-PAMI-07} and so on. We can roughly see these methods as a two-layer model though the parameters of local filters are hand-crafted. The first layer is usually applied on input image in a local and nonlinear way, such as Gabor magnitude and LBP coding. Lots of papers show that ``locally nonlinear + fully connected linear'' architecture is definitely better than ``fully connected linear'' architecture. Gabor+LBP+LDA proposed in \cite{Lei-TIP-2011} is even a three-layer model, which obtained good performance by carefully tuning.

As we know, more deeper models (\#layers $>$ 3) based on hand-crafted filters are rarely reported in the literature. Because the filters (or parameters) of each layer are usually designed independently by hand and the dynamics between layers are hard to handle by human observations. Therefore, learning the parameters of all layers from data is the best way out. With the successful applications of CNN in image classification, it becomes the mainstream in the field of face recognition rapidly. Before CNN becomes popular, many good filter learning methods~\cite{Maturana-ACCV-2011,Lei-PAMI-2013} were proposed to learn the parameters of two-layer models but they are difficult to generalize to deep architectures. Following the current trend, this paper uses deep CNN to learn face representation from large scale dataset.

\section{CASIA-WebFace Dataset}
\label{sec:db}

\subsection{Name and Image Collection}

Dataset for face recognition only need two kinds of data: face image and identity. Randomly crawling face images from Internet and annotating them is nearly an impossible mission. IMDb is a well structured website containing rich information of celebrities, such as name, gender, birthday and photos. We first search the celebrities born between 1940 to 2014 year on the website, and then crawl the names of them.

\begin{figure}
  \centering
  \includegraphics[width=0.3\textwidth]{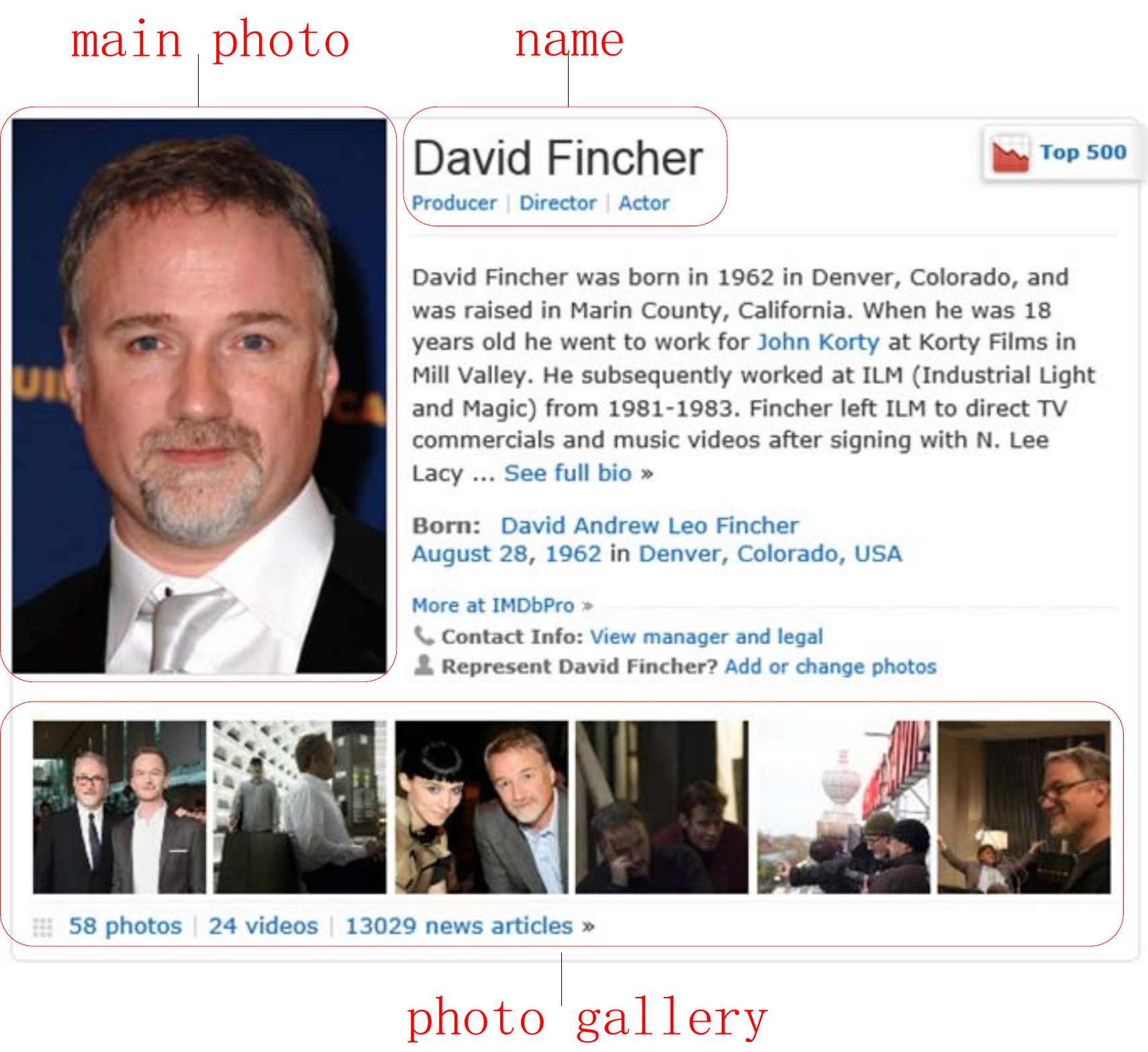}\\
  \caption{A sample page of David Fincher on IMDb. The ``main photo'' is used as initial seed and the 58 photos in the ``photo gallery'' need to be annotated.}
  \label{fig:celeb-page}
\end{figure}

Each celebrity has an independent page on the website. A sample page is shown in Fig.~\ref{fig:celeb-page}, in which we only focus on the ``name'', ``main photo'' and ``photo gallery'' contents. Neglecting the celebrities don't having ``main photo'', we get 38,423 subjects and 903,304 images in total. Then all images are processed by a multi-view face detector, 844,126 images remain in the dataset and 1,556,352 faces are detected. Because many images appear in the ``photo gallery'' of serval celebrities simultaneously, the actual number of images and faces are smaller than above numbers.

\subsection{Face Annotation}

The dataset at current state can't be used for training, we need annotate the identity of faces in each image. The ``main photo'' usually only contains a single face of the corresponding celebrity but the majority of photos in the ``photo gallery'' contain multiple faces which belong to other celebrities. Our task is to assign an identity to each face, and divide the faces into groups according to their identities.

\begin{figure}
  \centering
  \includegraphics[width=0.5\textwidth]{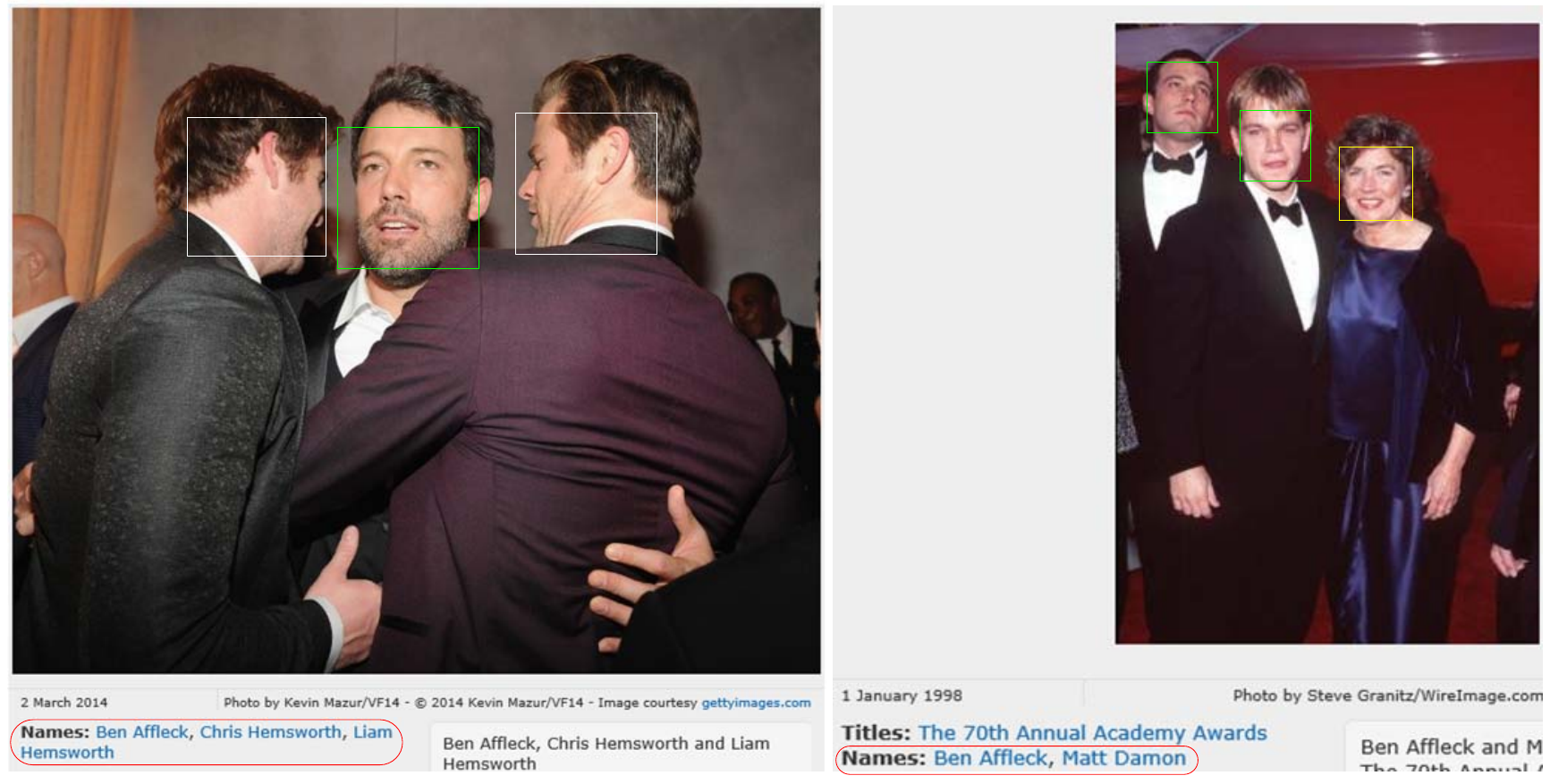}\\
  \caption{Two sample photos of Ben Affleck containing multiple faces. The name tags corresponded to the photo are shown at the left-bottom of photo. The left photo contains 3 faces and is corresponded to 3 names, but 2 faces are not detected (white rectangles). The right photo contains 3 faces but is only corresponded to 2 names. The woman in the right figure is not annotated (yellow rectangle).}
  \label{fig:celeb-photo}
\end{figure}

Browsing the ``photo gallery'' of each celebrity, we find that every photo is annotated by several name tags. The name tag can reduce the search space of face-identity correspondence and simply our annotation task. Two sample pages of photo are shown in Fig.~\ref{fig:celeb-photo}, which illustrates two kinds of noise in the photo page: miss detection and miss annotation.

Clustering all faces by existing face recognition engine is a natural way to deal this large scale task. General clustering methods need to compute the similarity (or distance) matrix of all samples first but the matrix is too large to be loaded into memory. To effectively annotate the large scale faces, we propose a three-step way by using name tag and face similarity simultaneously. The proposed method can run on normal PC and obtain good clustering results. The steps of our tag-similarity clustering method are as follows.
\begin{enumerate}
\item Extract the feature template of each face by a pre-trained face recognition engine~\cite{Yi-CVPR-2013};
\item Use the ``main photo'' of each celebrity as its seed.
\item Use the images containing 1 face to augment each celebrity's seeding images.
\item For the remain images in ``photo gallery'', find the correspondence between faces and celebrities constrained by similarity and name tag.
\item Crop faces from images and save into independent folder for each celebrity. Manually check the dataset and delete the false grouped face images.
\end{enumerate}

After clustering complete, we remove the subjects having less than 15 face images. To make this dataset compatible with LFW, we check the duplicate subjects based on edit distance between the names in CASIA-WebFace and LFW. There are 1043 subjects with the same names are found between CASIA-WebFace and LFW, and these subjects are removed from CASIA-WebFace. Now, we can see CASIA-WebFace as an independent training set for LFW. By combining CASIA-WebFace and LFW, we obtain a new benchmark for large scale face recognition in the wild.

After being cleaned, CASIA-WebFace finally has 10,575 subjects and 494,414 face images. Because the scale of dataset is too large, we can't promise all faces are detected and annotated correctly. A small amount of miss classified samples don't affect the training process and may be able to improve the robustness of the model. The quality of the dataset will be illustrated by the experimental results.

\subsection{Dataset Statistics}

The statistics of the proposed CASIA-WebFace dataset is shown in Table~\ref{tbl:db-stat}. Except for Facebook's SFC dataset, the scale of CASIA-WebFace has the largest scale. For users' privacy issue, maybe SFC will never be open to research community. The features of Microsoft's WDRef dataset was publicly available from 2012 but it is inflexible for advanced researches. Among the datasets listed in the table, CASIA-IMDb+LFW is the most suitable combination for large scale face recognition in the wild. If you feel the accuracy of LFW has been saturated by the current state-of-the-art method~\cite{Sun-NIPS-2014}. BLUFR~\cite{Liao-IJCB-2014} is a more challenging protocol to report your results.

\begin{table*}
  \centering
  \begin{tabular}{|l|l|l|l|}
    \hline
    Dataset & \#Subjects & \#Images & Availability \\
    \hline
    \hline
    LFW~\cite{Huang-LFW-2007} & 5,749 & 13,233 & Public \\
    \hline
    WDRef~\cite{Chen-ECCV-2012} & 2,995 & 99,773 & Public (feature only) \\
    \hline
    CelebFaces~\cite{Sun-NIPS-2014} & 10,177 & 202,599 & Private \\
    \hline
    SFC~\cite{Taigman-CVPR-2014} & 4030 & 4,400,000 & Private \\
    \hline
    CACD~\cite{Chen-ECCV-2014} & 2,000 & 163,446 & Public (partial annotated) \\
    \hline
    CASIA-WebFace & 10,575 & 494,414 & Public \\
    \hline
  \end{tabular}
  \caption{The information of CASIA-WebFace and comparison to other large scale face datasets.}
  \label{tbl:db-stat}
\end{table*}

\section{Learning Deep Representation}
\label{sec:learning}

To illustrate the advantages of the proposed dataset, we learn an effective representation from this dataset by deep convolutional network with many latest tricks.

\subsection{Convolutional Network}

The baseline deep convolutional network is constructed by combining many tricks from recent successful networks including very deep architecture~\cite{Simonyan-2014}, low dimensional representation and multiple loss functions~\cite{Sun-NIPS-2014}. Small filter and very deep architecture can reduce the number of parameters and enhance the nonlinearity of the network. Low dimensional representation conforms to the assumption that face images usually lie on a low dimensional manifold and the low dimensional constrain can reduce the complexity of the network. Combing identification and verification loss functions has been analyzed in \cite{Sun-NIPS-2014}, which can learn more discriminative representations than Softmax only.

The dimension of input layer is 100$\times$100$\times$1 channel, \ie gray image. The proposed network includes 10 convolutional layer, 5 pooling layers and 1 fully connected layers, the detailed architecture of which is shown in Table~\ref{tbl:arch}. The size of all filters in the network are 3$\times$3. The first four pooling layers use max operator and the last pooling layer is average. Limited by the computation power of our GPU, the architecture is not optimal but just determined according to our experience. There are still some room to improve.

\begin{table*}
  \centering
  \begin{tabular}{|l|l|c|c|c|c|}
    \hline
    Name & Type & \tabincell{c}{Filter Size\\/Stride} & Output size & Depth & \#Params\\
    \hline
    \hline
    Conv11 & convolution & 3$\times$3 / 1 & 100$\times$100$\times$32 & 1 & 0.28K \\
    Conv12 & convolution & 3$\times$3 / 1 & 100$\times$100$\times$64 & 1 & 18K \\
    \hline
    Pool1 & max pooling & 2$\times$2 / 2 & 50$\times$50$\times$64 & 0 & \\
    \hline
    Conv21 & convolution & 3$\times$3 / 1 & 50$\times$50$\times$64 & 1 & 36K \\
    Conv22 & convolution & 3$\times$3 / 1 & 50$\times$50$\times$128 & 1 & 72K \\
    \hline
    Pool2 & max pooling & 2$\times$2 / 2 & 25$\times$25$\times$128 & & \\
    \hline
    Conv31 & convolution & 3$\times$3 / 1 & 25$\times$25$\times$96 & 1 & 108K \\
    Conv32 & convolution & 3$\times$3 / 1 & 25$\times$25$\times$192 & 1 & 162K \\
    \hline
    Pool3 & max pooling & 2$\times$2 / 2 & 13$\times$13$\times$192 & 0 & \\
    \hline
    Conv41 & convolution & 3$\times$3 / 1 & 13$\times$13$\times$128 & 1 & 216K \\
    Conv42 & convolution & 3$\times$3 / 1 & 13$\times$13$\times$256 & 1 & 288K \\
    \hline
    Pool4 & max pooling & 2$\times$2 / 2 & 7$\times$7$\times$256 & 0 & \\
    \hline
    Conv51 & convolution & 3$\times$3 / 1 & 7$\times$7$\times$160 & 1 & 360K \\
    Conv52 & convolution & 3$\times$3 / 1 & 7$\times$7$\times$320 & 1 & 450K \\
    \hline
    \bf{Pool5} & avg pooling & 7$\times$7 / 1 & 1$\times$1$\times$320 & & \\
    \hline
    Dropout & dropout (40\%) & & 1$\times$1$\times$320 & 0 & \\
    \hline
    Fc6 & fully connection & & 10575 & 1 & 3305K \\
    \hline
    Cost1 & softmax & & 10575 & 0 & \\
    \hline
    Cost2 & contrastive & & 1 & 0 & \\
    \hline
    \hline
    Total & & & & 11 & 5015K \\
    \hline
  \end{tabular}
  \caption{The architecture of the proposed baseline convolutional network.}
  \label{tbl:arch}
\end{table*}

Small filter and very deep architecture were independently proposed in \cite{Simonyan-2014} and \cite{Szegedy-2014}. \cite{Simonyan-2014} achieved high performance in ImageNet 2014 challenges by a 19 layer network. Meanwhile, \cite{Szegedy-2014} obtained slightly better results than \cite{Simonyan-2014} by a 22 layer network. This paper combines the tricks from these two papers. We use multiple small filters to approximate large filter and remove redundant fully connected layers to reduce the number of parameters. Finally, our network uses 3$\times$3 filter in all 10 convolutional layers and just has 1 fully connected layer.

Pool5 layer is used as face representation, the dimension of which is equal to the number of channel of Conv52, 320. To distinguish large number of subjects in the training set (10575), this low dimensional representation should fully distill discriminative information from face images. As same as \cite{Sun-NIPS-2014}, Softmax (identification) and Contrastive (verification) cost are combined to construct the objective function. ReLU neuron is used after all convolutional layers, except for Conv52. Because Conv52 are combined by average to generate the low dimensional face representation, they should be dense and compact. ReLU is apt to produce sparse vector, therefore applying it on the face representation will degrade the performance.

In the training stage, Pool5 is used as input of Contrastive cost function. And Fc6 is used as input of Softmax cost function. Because the number of parameters of Fc6 is very large, \ie 320$\times$10575, we set the dropout ratio as 0.4 to regularize Fc6. The importance of two cost functions are balanced by a weight $\alpha$.

\begin{figure}
  \centering
  \includegraphics[width=0.5\textwidth]{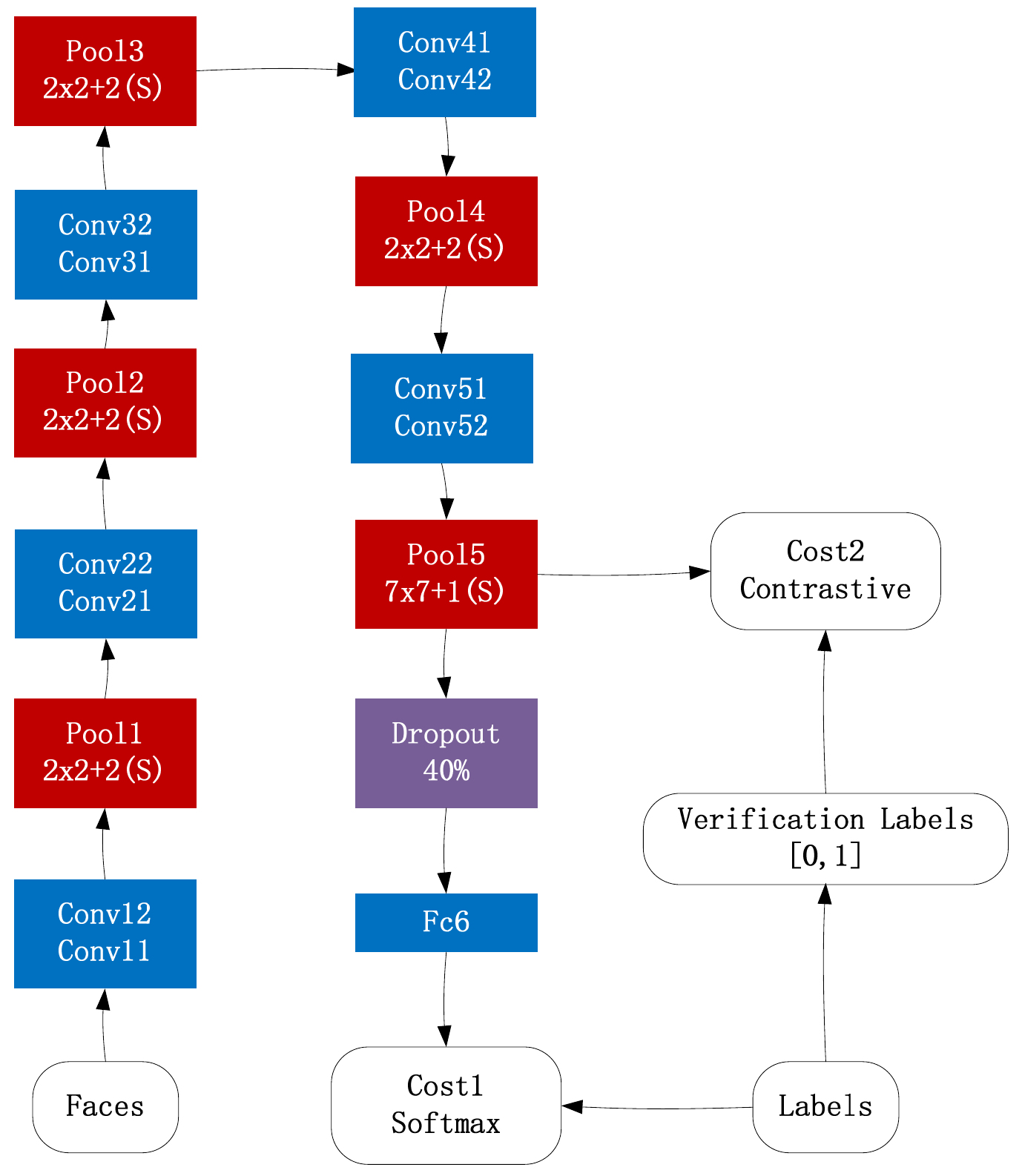}\\
  \caption{The proposed baseline convolutional network with many recent tricks.}
  \label{fig:net}
\end{figure}

\subsection{Training Methodology}
\label{sec:train-steps}

\begin{figure}
  \centering
  \includegraphics[width=0.3\textwidth]{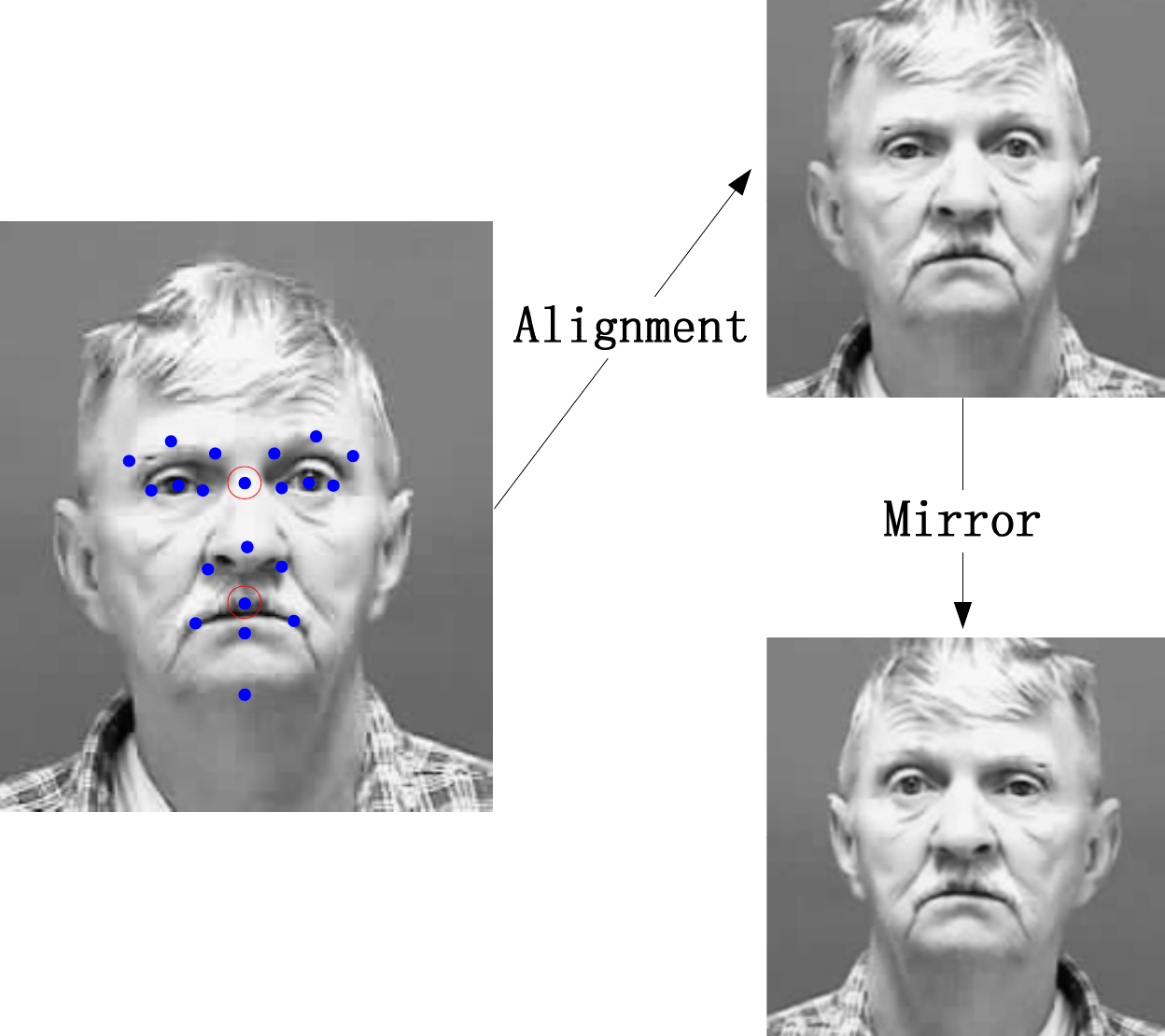}\\
  \caption{Face image alignment and augmentation. The read circles on the face are two selected landmarks for similarity transformation.}
  \label{fig:alignment}
\end{figure}

Before input to the network, all face images are converted to gray scale and normalized to 100$\times$100 according to two landmarks (see Fig.~\ref{fig:alignment}). Compared to the most used eye centers, the distance between the selected two landmarks here is relative invariant to pose variations in yaw angle. After normalization, the distance between the two points is 25 pixels. Because face has nearly symmetric structure, we double the training set by mirror operation, which can result the representations more robust to pose variations.

 With such huge amounts of data, the current network is unlikely over-fitting, therefore we set the weight decay of all convolutional layers to 0 and the weight decay of the fully connected layer to 5e-4. The learn rate is set to 1e-2 initially and reduce to 1e-5 gradually. Because the convergence rate of Softmax is faster than Contrastive cost function, the weight $\alpha$ is set to a small value 3.2e-4 at first and increase to 6.4e-3 gradually.

The open source implementation cuda-convnet~\cite{cuda-convnet} is used to train our network. For Softmax cost, we just need input face images and their labels, but for Contrastive cost, we need generate face pairs by sampling from the training set. To reduce the consumption of memory and disk space, we just sample the positive and negative face pairs online within each batch. The face pairs across batch are not covered. How to generate complete face pairs effectively is left to future work.

\section{Experiments}
\label{sec:exp}

CASIA-WebFace is always used to train the deep network. As described in the last section, all face images in CASIA-WebFace are processed by face detection, face landmarking and alignment. Due to the facial symmetry, we mirror 493,456 detected faces to augment the dataset. Finally, we have 986,912 training samples. The whole process is fully automatic and the false aligned faces are remained as they are. There are still a small number of miss detection in this dataset. If you have a better face detector than ours, your training set may be larger than ours slightly.

LFW and YTF are two most popular and challenging datasets for face recognition in the wild. Because they are not overlapped to the proposed CASIA-WebFace, it's very reasonable to report performance on LFW and YTF. Besides that, the trained deep network is also evaluated according to a more challenging and practical protocol, BLUFR, which can reflect the performance of face recognition in real applications more objectively.

\subsection{Results on LFW}

LFW includes 5,749 subjects and 13,233 face images. There are three main protocols for performance reporting: unsupervised, restricted and unrestricted protocol. Unsupervised protocol is used to evaluate the baseline performance of face representation and the other two protocols are usually used to evaluated the performance of metric learning or the whole method. For all protocols, the test set is fixed, which includes 6000 face pairs in 10 splits. Mean accuracy and standard error of the mean should be reported.

\subsubsection{Standard Protocol}

All images in LFW are processed by the same pipeline as CASIA-WebFace, and normalized to $100\times100$. The representations of all faces are extracted by the deep network trained on CASIA-WebFace (be abbreviated to DR). First, we evaluate the base performance of the representation directly. Then, we test the influence of unsupervised and supervised learning on the base representations. The following experiments are conducted:
\begin{itemize}
\item A: DR + Cosine;
\item B: DR + PCA on CASIA-WebFace + Cosine;
\item C: DR + Joint Bayes on CASIA-WebFace;
\item D: DR + PCA on LFW training set + Cosine;
\item E: DR + Joint Bayse on LFW training set.
\end{itemize}

\begin{table}
  \centering
  \begin{tabular}{|l|l|l|l|}
    \hline
    Method & \#Net & Accuracy$\pm$SE & Protocol \\
    \hline
    DeepFace & 1 & $95.92\pm0.29\%$ & unsupervised \\
    DeepFace & 1 & $97.00\pm0.28\%$ & restricted \\
    DeepFace & 3 & $97.15\pm0.27\%$ & restricted \\
    DeepFace & 7 & $97.35\pm0.25\%$ & unrestricted \\
    \hline
    \hline
    DeepID2 & 1 & $95.43\%$ & unrestricted \\
    DeepID2 & 2 & $97.28\%$ & unrestricted \\
    DeepID2 & 4 & $97.75\%$ & unrestricted \\
    DeepID2 & 25 & $98.97\%$ & unrestricted \\
    \hline
    \hline
    Ours A & 1 & $96.13\pm0.30\%$ & unsupervised \\
    Ours B & 1 & $96.30\pm0.35\%$ & unsupervised \\
    Ours C & 1 & $97.30\pm0.31\%$ & unsupervised~\tablefootnote{The term ``unsupervised'' means that the model is not trained on LFW in supervised way.} \\
    Ours D & 1 & $96.33\pm0.42\%$ & unsupervised \\
    Ours E & 1 & $97.73\pm0.31\%$ & unrestricted \\
    \hline
  \end{tabular}
  \caption{The performance of our baseline deep networks and compared methods on LFW View2.}
  \label{tbl:lfw}
\end{table}

According the protocols of LFW, the hyper-parameters are both tuned on CASIA-WebFace or View1 of LFW, such as the dimension of PCA and the regularization factor of Joint Bayes. The the accuracies are evaluated on View2 of LFW and listed in Table~\ref{tbl:lfw}. Other state-of-the-art results of DeepFace and DeepID2 are also given for comparison. From the results of our 6 experiments, we can draw 4 conclusions:
\begin{enumerate}
\item The base representation has good performance;
\item Fine-tuning on the training set of LFW can improve the performance slightly, \eg B$\rightarrow$D, C$\rightarrow$E;
\item Based on the base representation, Joint Bayes can improve the performance marginally, \eg B$\rightarrow$C, D$\rightarrow$E.
\end{enumerate}

By inspecting the results in unsupervised setting, we can see that our base representation is better than DeepFace, 96.13\% vs. 95.92\%. After tuning on LFW by PCA, the accuracy 96.33\% is improved slightly. Because Joint Bayes can't deal with pairwise samples directly, we don't conduct experiment by restricted protocol. When using unrestricted protocol, our single-network scheme E achieves 97.73\%, which is better than DeepFace's 7-networks ensemble 97.35\% and is comparable to DeepID2's 4-networks ensemble 97.75\%.

The superiority of our network is benefit from the deep architecture. On other aspects, our method and dataset are still inferior to DeepFace: 1) we just align face images by 2D similarity transformation which is inferior to DeepFace's 3D alignment; 2) The scale of training set of DeepFace, SFC, is 10X larger than our CASIA-WebFace. Limited by GPU computational resources, we don't continue to train deeper network or train network ensemble to improve the performance here. After publish the dataset, CASIA-WebFace, we believe the whole research community can refresh the record more quickly.

\subsubsection{BLUFR Protocol}

The test set of LFW just include 6000 face pairs, half of which is genuine and the other half is impostor. As discussed in \cite{Liao-IJCB-2014}, such scale of negative face pairs are not enough to evaluate the performance at low FARs. Therefore, \cite{Liao-IJCB-2014} developed a new benchmark protocol, called BLUFR, to fully use all the 13,233 face images in LFW. BLUFR contains both verification and open-set identification scenarios, with a focus at low FARs. There are 10 trials of experiments, with each trial containing about 156,915 genuine matching scores and 46,960,863 impostor matching scores on average for performance evaluation.

The representations of faces in LFW are extracted in the same way as the previous experiment. Then the results can be calculated by the standard benchmark toolkit~\cite{blufr}. For simplicity, we just report the results of scheme E and F. The VR (Verification Rate) and DIR (Detection and Identification Rate) of our methods and compared methods are listed in Table~\ref{tbl:blufr}. The numbers in the table are measured in $(\mu- \sigma)$ of 10 trials, where $\mu$ is mean accuracy and $\sigma$ is standard deviation.

\begin{table}
  \centering
  \begin{tabular}{|l|l|l|}
    \hline
    Method & VR@FAR=0.1\% & \tabincell{c}{DIR@FAR=1\%\\Rank=1} \\
    \hline
    HD-LBP + JB & $41.66\%$ & $18.07\%$ \\
    HD-LBP + LDA & $36.12\%$ & $14.94\%$ \\
    \hline
    \hline
    Ours E & $80.26\%$ & $28.90\%$ \\
    \hline
  \end{tabular}
  \caption{The performance of our baseline deep network and compared methods on LFW under BLUFR protocol.}
  \label{tbl:blufr}
\end{table}

\cite{Liao-IJCB-2014} just reported the performance of some conventional shallow (but wide) models under BLUFR protocol. The best reported method is HD-LBP + JB (High Dimensional LBP + Joint Bayes), the result of which is VR=41.66\% (at FAR=0.1\%). As shown in Table~\ref{tbl:blufr}, our deep network surpass HD-LBP based methods significantly. The superiority of deep models against wide models has been illustrated in previous work and the conclusion is verified in this paper again.

We find that all numbers in Table~\ref{tbl:blufr} are obviously lower than those in Table~\ref{tbl:lfw}, especially the DIR (at FAR=1\% and Rank=1). Because DIR is an important index to reflect the performance of face surveillance (or watch-list) systems, we think face recognition algorithms still have large gap to appeal the requirements of surveillance applications.

\subsection{Results on YTF}

To test the generalization ability of our network, we evaluate it on a video face dataset, YouTube Faces (YTF). Due to motion blur and high compression ratio, the quality of images in YTF are much worse than web photos. For each video in YTF, we randomly select 15 frames and extract their representations by our deep network (DR). In the training stage, the 15 frames are seen as 15 samples with same identity. In the testing stage, the similarity score of video pair is the mean value of 15$\times$15=225 frame pairs. The following experiments are conducted in unsupervised and supervised settings:
\begin{itemize}
\item A: DR + Cosine;
\item D: DR + PCA on YTF training set + Cosine;
\item E: DR + Joint Bayes on YTF training set.
\end{itemize}

\begin{table}
  \centering
  \begin{tabular}{|l|l|l|l|}
    \hline
    Method & \#Net & Accuracy & Protocol\\
    \hline
    DeepFace & 1 & $91.4\pm{1.1}\%$ & supervised \\
    \hline
    \hline
    Ours A & 1 & $88.00\pm{1.50}\%$ & unsupervised \\
    Ours D & 1 & $90.60\pm{1.24}\%$ & unsupervised \\
    Ours E & 1 & $92.24\pm{1.28}\%$ & supervised \\
    \hline
  \end{tabular}
  \caption{The performance of our methods and DeepFace on Youtube Faces (YTF).}
  \label{tbl:ytf}
\end{table}

DeepFace holds the best result on YTF and is better than other methods by a large margin, therefore, we only compare to DeepFace. Directly matching by Cosine function, the base representation achieves 88.00\% accuracy on YTF. Transforming the representation by PCA on YTF, the accuracy improves to 90.60\% remarkably. When tuning the representation by Joint Bayes further, our method outperforms DeepFace slightly.

\section{Conclusion}
\label{sec:con}



This work collected a large scale face dataset from Internet and made it public to research community. The new dataset isn't overlapping to LFW and can be used in conjunction with LFW for large scale face recognition research. This combination can standardize the evaluation protocol of LFW and advance the reproducible research. On the other side, unified training and testing set can make various methods comparable. This work also described the whole process of dataset construction and face representation learning by a 11-layer convolutional network. Referring the pipeline proposed in this paper, anyone can easily train a high performance face recognition engine. Future work will be done in three directions: 1) augment the dataset by using commercial image search engines; 2) develop more effective annotation tools and algorithms; 3) explore novel methods to train single network to approach the performance of big deep network ensemble.

\section*{Acknowledgements}

This work was supported by the Chinese National Natural Science Foundation Projects \#61105023, \#61103156, \#61105037, \#61203267, \#61375037, \#61473291‍, National Science and Technology Support Program Project \#2013BAK02B01, Chinese Academy of Sciences Project No.KGZD-EW-102-2, and AuthenMetric R\&D Funds.

The Tesla K40 used for this research was donated by the NVIDIA Corporation.

{\small
\bibliographystyle{ieee}
\bibliography{vision}
}

\end{document}